%% file: sigconf.tex
\documentclass[sigconf]{acmart}

\usepackage{booktabs} 
\usepackage{graphicx}
\usepackage{tabularx}
\newcolumntype{Y}{>{\centering\arraybackslash}X}

\setcopyright{rightsretained}

\acmDOI{10.1145/nnnnnnn.nnnnnnn}

\acmISBN{978-x-xxxx-xxxx-x/YY/MM}

\acmConference[GECCO '21]{the Genetic and Evolutionary Computation Conference 2021}{July 10--14, 2021}{Lille, France}
\acmYear{2021}
\copyrightyear{2021}

\acmPrice{15.00}
\copyrightyear{2021}
\acmYear{2021}
\setcopyright{acmcopyright}\acmConference[GECCO '21 Companion]{2021 Genetic and Evolutionary Computation Conference Companion}{July 10--14, 2021}{Lille, France}
\acmBooktitle{2021 Genetic and Evolutionary Computation Conference Companion (GECCO '21 Companion), July 10--14, 2021, Lille, France}
\acmPrice{15.00}
\acmDOI{10.1145/3449726.3463221}
\acmISBN{978-1-4503-8351-6/21/07}

\begin{document}
\title{Concurrent Neural Tree and Data Preprocessing AutoML for Image Classification}
\author{Anish Thite}
\author{Mohan Dodda}
\author{Alex Liu}
\author{Pulak Agarwal}
\affiliation{\institution{Georgia Institute of Technology}}
\author{Jason Zutty}
\affiliation{\institution{Georgia Tech Research Institute}}

\begin{abstract}
Deep Neural Networks (DNN's) are a widely-used solution for a variety of machine learning problems. However, it is often necessary to invest a significant amount of a data scientist's time to pre-process input data, test different neural network architectures, and tune hyper-parameters for optimal performance. Automated machine learning (autoML) methods automatically search the architecture and hyper-parameter space for optimal neural networks. However, current state-of-the-art (SOTA) methods do not include traditional methods for manipulating input data as part of the algorithmic search space. We adapt the Evolutionary Multi-objective Algorithm Design Engine (EMADE), a multi-objective evolutionary search framework for traditional machine learning methods, to perform neural architecture search. We also integrate EMADE's signal processing and image processing primitives. These primitives allow EMADE to manipulate input data before ingestion into the simultaneously evolved DNN. We show that including these methods as part of the search space shows potential to provide benefits to performance on the CIFAR-10 image classification benchmark dataset.
\end{abstract}

%
%

\ccsdesc[500]{Computing methodologies~Computer vision tasks}
\ccsdesc[500]{Computing methodologies~Search methodologies}
\ccsdesc[500]{Computing methodologies~Genetic programming}
\ccsdesc[500]{Computing methodologies~Neural networks}

\keywords{Deep Neural Networks (DNNs), Automated Machine Learning (autoML), Genetic Programming}

\maketitle

\input{main_body}

\bibliographystyle{ACM-Reference-Format}
\bibliography{sample.bib} 

\end{document}

%% file: main_body.tex
\section{Introduction}



Deep Learning methods have been shown to improve state-of-the-art performance in a variety of tasks, including Natural Language Processing, Computer Vision, and Speech Processing. This has led to a flurry of activity in the field, with many different task-specific model architectures and hyper-parameters being invented. 
Due to the wide variety of possible architectures, hyper-parameters, it is often necessary to spend copious amounts of time tracking and experimenting with many different architectures and hyper-parameters for optimal performance. 
Neural Architecture Search (NAS) methods help automate this process. 
Methods treat finding the optimal architecture as a search problem and employ a variety of strategies in order to solve it. 
Hyper-parameter search methodologies also exist, and modern systems incorporate both hyper-parameters and architectures in the same search space. 


Optimal performance can be measured with single or multiple objectives. For example, LEAF \cite{leaf}, a multi-objective system optimizes for both accuracy and model size. Since each algorithm is tested on both objectives, end users can select the ideal model based on their desired compute budget. 

One major drawback with current systems is a lack of data pre-possessing in the search space. 
Rather, data pre-processing is hand-designed prior to feeding it into the system. 
Data pre-processing is arguably one of the most important pieces of a deep learning pipeline, and can dramatically affect the quality of the resulting model.

The Evolutionary Multi-objective Algorithm Design Engine (EM\-ADE) tackles this problem. EMADE, previously introduced as GTMOEP \cite{zutty2015multiple}, includes data processing functions in its primitive set, allowing it to evolve both traditional machine learners and data pre-processing algorithms concurrently.
Using the Keras API \cite{keras} we extend EMADE to include neural network-relevant primitives and terminals such as various types of layers, weight initializers, layer-level parameters, and network-level hyper-parameters. 
We show that modifying EMADE to evolve using both data pre-processing functions and neural network primitives improves accuracy on CIFAR-10\cite{cifar}, a popular benchmark for NAS frameworks, and can potentially improve on SOTA algorithms.

\section{Background and Related Work}


Neural Architecture Search automates the process of determining the best combinations of layers and their optimal hyper-parameters for neural networks by treating it as a classical search problem. NAS methods comprise three components: a search space, a search strategy, and a candidate network performance estimation strategy. 

Evolution-based search algorithms have been used widely to evolve neural networks. 
NEAT \cite{stanley2002evolving} was one of the first genetic algorithms for neural network topology evolution, created by Stanley \& Miikkulainen in 2002. 
NEAT directly encodes neurons, their weights, and neuron connections as genomes. Then NEAT creates a population of genomes which then undergoes a biologically-inspired evolutionary process. 
Genomes can mutate (where the weights and connection values randomly change) and mate (where portions of two genomes combine to make a new one).

However, direct encoding of neurons to genes is not feasible at larger scales, due to the amount of memory required to store millions of neurons. CoDEEPNEAT \cite{miikkulainen2019evolving} and HyperNEAT \cite{hyperneat} encode neural networks indirectly by specifying layer parameters and sizes rather than individual neurons. This allows NEAT to be used to evolve larger and deeper neural networks. CoDEEPNEAT also performs multi-objective optimization, resulting in a list of co-dominant networks (called a Pareto front). This allows CoDEEPNEAT to optimize for metrics such as accuracy and size simultaneously. Experiments on CoDEEPNEAT were run on a variety of datasets \cite{leaf} including CIFAR-10 \cite{miikkulainen2019evolving}. LEAF creates networks from scratch and trains them for 8 epochs during the evolution process. Once the evolution process is complete, the best network is re-trained for 300 epochs. LEAF achieves a 7.3\% error rate on CIFAR-10. 


EMADE is a automated ML framework based on the DEAP \cite{DEAP_JMLR2012} genetic programming framework. In the DEAP framework, algorithms are expressed as trees, with functions expressed as nodes and parameters expressed as terminals. When evaluating the tree, nodes are computed in order of decreasing depth with outputs from child nodes being passed as inputs into their parent nodes. EMADE uses vector based genetic programming, where the data passed between primitives is a vector of elements. EMADE uses high-level primitives such signal processing and computer vision functions along with traditional machine learning functions.


Through the evolutionary process EMADE optimizes combinations of these classifiers and pre-processing methods and their respective hyper-parameters. EMADE's evolutionary process follows a similar algorithm to NEAT and other genetic algorithms. An initial population of individual trees are created and evaluated. Each generation, $n$ trees are selected via a selection function, and are used to create new trees to add to the population. EMADE uses NSGA-II \cite{deb2000fast} as the selection function, and a variety of mating and mutation operators for tree generation. EMADE has been used for a variety of domains \cite{griffies2018wearable, zutty2019reducing, zutty2018emade}, but has never been used on Deep Learning applications. 

\section{Methods}

\subsection{Neural Network Modifications}
We extend EMADE to include a neural network-based learner primitive, Layer primitives, and neural network-centered mating and mutation functions. 

\subsubsection{LayerTree}

The LayerTree class is a pre-ordered list representation of the neural network tree. Primitives add Keras layers to this tree, resulting in a tree-structured computational graph.

\subsubsection{Layer Primitives}
It is necessary to encode the neural architecture in a way visible to the EMADE framework. Therefore, a multitude of neural network layer types from the Keras API are added as EMADE primitives. 
Keras Layers are added as Layer primitives. Each primitive accepts a LayerTree representation of the model built from its child node as a parameter. It creates a Keras layer and adds the layer as the new root, and passes the modified tree on to its parent node. In creating the Keras Layer, EMADE can specify various layer parameters via other primitives and terminals. These parameters are shown in \autoref{tab:hyperparams}. 

We include Dense and Convolution Layers in the primitive set since CIFAR-10 is an image classification task, and CNNs have performed well on this task.
We also add layer primitives including Dropout, BatchNormalization, MaxPooling, GlobalMaxPooling, and GlobalAveragePooling. Concatenate layers are also necessary for EMADE to generate tree-based neural networks. These concatenate a series of previous layers' outputs, and can be used to condense two branches of the tree together. Therefore, we add a Concatenate layer primitive to the set. The primitive takes the output of two child LayerTrees and concatenates them. 

\subsubsection{Pre-trained Layer Primitives}
Many state-of-the-art image classification models show that pre-training on ImageNet yields significant performance gains \cite{dosovitskiy2020image} \cite{wu2021cvt} \cite{tan2021efficientnetv2}. Building off this work, we incorporated pre-trained models within the evolutionary process. We add each pre-trained network as a single layer. While it does not allow EMADE to modify the architecture of the pre-trained network, EMADE is easily able to add other primitives before or after these layers.
We implement VGG \cite{vgg}, Mobilenet \cite{mobilenet}, and Inception \cite{inception} architectures, which have been shown to perform well for image classification tasks. For all of these layers, we used pre-trained ImageNet weights. However, we do not freeze these weights, allowing them to be updated during the training regime for each tree.


\subsubsection{NNLearner}
The NNLearner takes in data, batch size, optimizer, and a LayerTree. It creates a neural network from the LayerTree specifications using the Keras Functional API. When evaluating, it trains the network on the train subset of the data, and validates it based on the validation set of the data. NNLearner uses early stopping \cite{earlystopping} to determine the optimal epoch to stop training using validation data to determine the early stopping point. 
\begin{table}[t]
\tiny
\caption{Hyper-parameters for Search Space}
\begin{tabular}{@{}lll@{}}
\toprule
 Hyper-parameter & Used In & Range \\ \midrule
Output Dimension & Core-Layer & int \\
Activation Function & Core-Layer & [SeLU, ELU, Sigmoid, ReLU, Leaky Relu, Softmax, TanH] \\ 
Weight Decay & Core-Layer & float \\
Kernel Dimension & CNN & int \\
Stride Size & CNN & int \\
Padding Style & CNN & Same or Valid \\
Batch Size & NNLearner & int \\
Optmizer & NNLearner & [Adam, SGD, RMSprop,  Adadelta, Adagrad, Adamax, Nadam, Ftrl] \\
Dropout Percentage & Dropout & float \\ \bottomrule
\end{tabular}
\label{tab:hyperparams}
\end{table}
\subsection{Image Pre-processing Primitives}
EMADE supports over 150 primitives that operate on image data, many leveraged from OpenCV. These include various filters, transformations, and other image processing techniques. EMADE also contains more than 60 signal processing functions leveraged from Python packages such as numpy, scipy, and scikit-learn.

Together, these primitives enable a large host of well-researched techniques to be applied to instances of data during the algorithm design process. Primitives in EMADE tend to fall into two categories, those that fit to training instances of data and then apply to those used for scoring (e.g. transforms), and those that apply to each instance independently (e.g. filters).

Some methods that we expect to be useful in pre-processing for image classification tasks include primitives such as: histogram of oriented gradients, hsv histograms, Haralick features, HuMoment features, edge detection filters, blurring filters, gray-scale conversions, normalizations, thresholds, and spectral representations such as Fourier transforms, wavelet transforms, and discrete cosine transforms. 

\subsection{Mutation Functions}
Various mutation functions were added to EMADE's mutation function set in order to aid neural network evolution. These mutation functions include randomly adding a layer to the tree representation, randomly swapping a layer with another layer, and randomly removing a layer. With these three simple mutation functions, EMADE has the potential to reach any architecture in the search space. Specific mutation functions were also added to change the activation function of a layer, the optimizer used, as well as the pre-trained weights used in the embedding layer. 

\subsection{Mating Functions}
\begin{figure}[h]
    \centering
    \includegraphics[width=4cm]{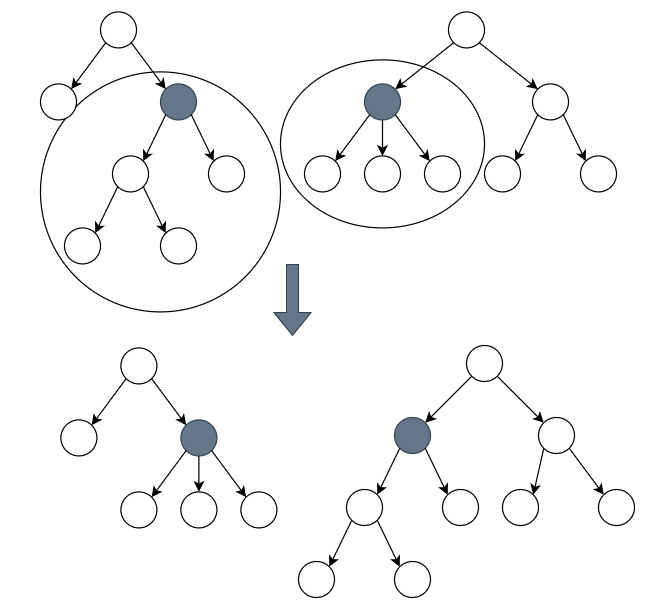}
    \caption{Single Point Crossover Between Two Trees}
    \label{fig:mating}
\end{figure}

Several new mating functions were added to EMADE's existing ones to improve mating between tree-based neural network algorithms. One mating function was a single point crossover that operates on a random point that has an NNLayer output type. This function allows the creation of a child network whose gene composition has its first partition of layers from the beginning of one parent up to the crossover point and the second partition of layers from the crossover point to the end of the other parent (and vice versa for the other child) as shown in \autoref{fig:mating}. A modified version of this function was also added to aid in expanding neural network size. Instead of replacing the latter half of each network, each child network has their original layers in addition to the swapped sub-trees from the parents. Note that this is not a two point crossover.

\section{Experiment}

\begin{table}[t]
\tiny
\caption{Experimental Setup}
\begin{center}
\begin{tabularx}{\columnwidth}{lYY}
\toprule
 Parameter & Small Run & Large Run \\ 
\midrule
Seeded Population Size & \multicolumn{2}{c}{16} \\
Initial Population Size (with seeded population) & 64 & 512 \\
Walltime & \multicolumn{2}{c}{24 hours } \\ 
Nodes & \multicolumn{2}{c}{4} \\ 
GPU / node & \multicolumn{2}{c}{1, 32GB TeslaV100} \\ 
Evaluations / node & \multicolumn{2}{c}{3} \\ 
Mating - crossover & \multicolumn{2}{c}{50\%} \\
Mating - crossoverEphemeral & \multicolumn{2}{c}{50\%} \\
Mating - headlessChicken & \multicolumn{2}{c}{10\%} \\
Mating - headlessChickenEphemeral & \multicolumn{2}{c}{10\%} \\
Mutation - insert & \multicolumn{2}{c}{5\%} \\
Mutation - insert modify & \multicolumn{2}{c}{10\%} \\
Mutation - ephemeral & \multicolumn{2}{c}{25\%} \\
Mutation - uniform & \multicolumn{2}{c}{5\%} \\
Mutation - shrink & \multicolumn{2}{c}{5\%} \\
Mutation - swap layer & \multicolumn{2}{c}{5\%} \\
Mutation - remove layer & \multicolumn{2}{c}{5\%} \\
Mutation - add layer & \multicolumn{2}{c}{5\%} \\
\bottomrule
\end{tabularx}
\label{table:expsetup}
\end{center}
\end{table}

The CIFAR-10 Dataset is a multi-class benchmark dataset for Image Classification. 
It contains 60,000 32x32 color images in 10 classes with 6,000 images per class. Classes consist of different vehicle and animal types. 
We separate the images into a 50,000 image train split and a 10,000 image test split. 

We initially seed EMADE's starting population with 16 hand-created NNLearners to bias the evolution process to create neural networks. Each seed is an NNLearner containing a pre-trained network layer, followed by a single Dense layer with a sigmoid activation function to act as an output layer. The pre-trained networks include either a VGG, MobileNet, or Inception architecture. Adam is set as the optimizer, and batch size differs between 1 to 9. The remaining individuals in the population are randomly generated. Using the configurations in \autoref{table:expsetup}, we performed two separate runs with the same seeded individuals but with different run sizes.  
Both runs were given approximately the same 288 total compute hours. The small run accomplished 109 generations and evaluated 13,896 individuals. The best individual accuracy was 75.59\%. The large run accomplished 28 generations and evaluated 34,989 individuals. The best individual accuracy was 73.84\%.



\section{Results}

\begin{figure}[h]    \centering
    \includegraphics[width=6cm]{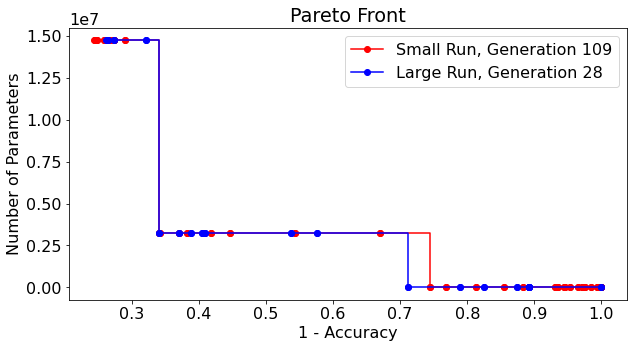}
    \caption{Final Pareto Fronts}
    \label{fig:pareto}
\end{figure}

\begin{figure}[h]
    \centering
    \includegraphics[width=6cm]{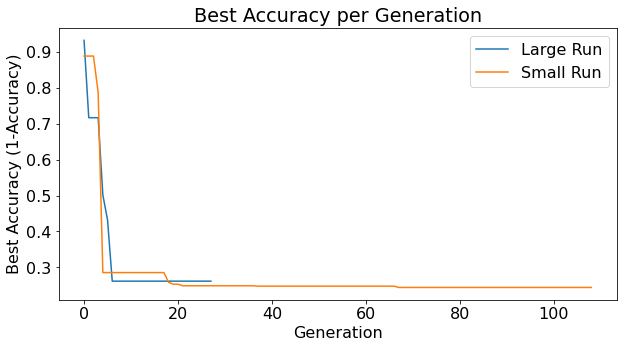}
    \caption{Best Accuracy per Generation}
    \label{fig:acc}
\end{figure}

Final Pareto Fronts are shown in \autoref{fig:pareto}. Even though the two runs performed a different number of generations and evaluated a significantly different amount of individuals the resulting Pareto Fronts are nearly identical. As seen in \autoref{fig:acc}, the best accuracy observed in an individual improved over the course of evolution. Note that the best individual accuracy improved the most in the early generations for both runs. Since both runs have the same accuracy/generation profiles, it seems that EMADE does not take advantage of the larger population to optimize more quickly. This similarity may be due to the effects of seeding, since the seeded portion of the starting population is the same. 

\begin{figure}[h]
    \centering
    \includegraphics[width=4.5cm]{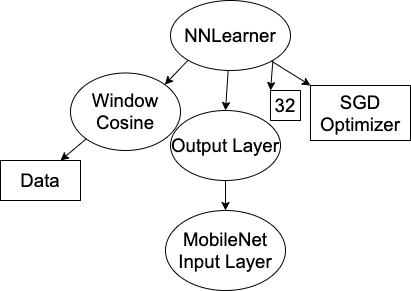}
    \caption{Individual with Pre-processing Primitive(s)}
    \label{fig:indpreproc}
\end{figure}

An example individual evolved is shown in \autoref{fig:indpreproc}. It takes advantage of both EMADE pre-processing and NN-based primitives. In this particular example, EMADE optimized the seeded pre-trained MobileNet network by manipulating the batch size, switching the optimizer from Adam to SGD, and applying a cosine window on the data before training/testing on it.


\section{Conclusion}
The paper shows a proof of concept that uses neuroevolution to look through a search space to find optimal neural network architectures across different objectives for image classification tasks. 
By evolving pre-trained networks and pre-processing techniques with the neural network architectures, we expand on current work to benefit the performance of the CIFAR-10 image classification dataset.
Although our current results are not yet comparable with human generated algorithms, it is worth noting that our algorithm was discovered by the EMADE neural network framework on its own using genetic programming. There is a lot of potential for machine generated algorithms to perform at an extremely high level.

We propose this tool to be helpful for other classification tasks where neural networks are used. There is also ample opportunity for future work in improving EMADE's neural architecture search, such as supporting skip connections within nodes in trees, investigating building and preserving tree complexity, and developing ways to more efficiently evaluate individuals. 

When EMADE's framework is further developed with more competitive results for the  CIFAR-10 dataset, we would like to test our results against other state-of-the-art algorithms and further investigate the benefits of manipulating input data as part of the algorithmic search space. Future research will also include more rigorous testing with other datasets (especially unbalanced ones) and drawing comparisons to other AutoML frameworks.

